\documentclass[twocolumn]{svjour3}
\usepackage[utf8]{inputenc}
\usepackage{graphicx}
\usepackage{amsmath}
\usepackage{multirow, makecell}
\usepackage{url}
\usepackage[section]{placeins}
\usepackage{verbatim}
\usepackage{tipa}

\DeclareMathOperator{\Att}{Att}
\DeclareMathOperator{\LSTM}{LSTM}
\DeclareMathOperator{\softmax}{softmax}

\DeclareMathOperator{\CNN}{CNN}

\title{Modelling word learning and recognition using visually grounded speech}

\author{Danny Merkx, Sebastiaan Scholten, Stefan L. Frank, Mirjam Ernestus and Odette Scharenborg}
\date{March 2022}
\institute{Corresponding author: D. Merkx \at
              Centre for Language Studies, Radboud University, the Netherlands \\
              Tel.: +31 24-3611461\\
              \email{danny.merkx@ru.nl}
           \and
           S. Scholten \at
              Multimedia Computing Group, Delft University of Technology, the Netherlands
            \and
            S. L. Frank \at
              Centre for Language Studies, Radboud University, the Netherlands
            \and
            M. Ernestus \at
              Centre for Language Studies, Radboud University, the Netherlands
            \and 
            O. Scharenborg \at
              Multimedia Computing Group, Delft University of Technology, the Netherlands
}

\begin{document}
\maketitle
\newpage

\begin{abstract}
\quad  \newline
\textbf{Background:} Computational models of speech recognition often assume that the set of target words is already given. This implies that these models do not learn to recognise speech from scratch without prior knowledge and explicit supervision. Visually grounded speech models learn to recognise speech without prior knowledge by exploiting statistical dependencies between spoken and visual input. While it has previously been shown that visually grounded speech models learn to recognise the presence of words in the input, we explicitly investigate such a model as a model of human speech recognition.

\textbf{Methods:}  We investigate the time-course of word recognition as simulated by the model using a gating paradigm to test whether its recognition is affected by well-known word-competition effects in human speech processing. We furthermore investigate whether vector quantisation, a technique for discrete representation learning, aids the model in the discovery and recognition of words. 

\textbf{Results/Conclusion:} Our experiments show that the model is able to recognise nouns in isolation and even learns to properly differentiate between plural and singular nouns. We also find that recognition is influenced by word competition from the word-initial cohort and neighbourhood density, mirroring word competition effects in human speech comprehension. Lastly, we find no evidence that vector quantisation is helpful in discovering and recognising words. Our gating experiments even show that the vector quantised model requires more of the input sequence for correct recognition. 

\end{abstract}
\noindent\textbf{Keywords}: computational modelling, speech recognition, multimodal learning, deep learning, vector quantisation
\section{Introduction}

Infants initially have little understanding of what is being said around them, and yet at approximately nine months old are able to produce their first words. When they start producing their first multi-word utterances around 18 months, they can already produce about 45 words and comprehend many more \cite{benedict,bates}. One of the challenges infants face is that speech does not contain neat breaks between words, which would allow them to segment the utterance into words. To complicate things further, words might be embedded in longer words (e.g., \emph{ham} in \emph{hamster}) and furthermore, no two realisations of the same spoken word are ever the same due to speaker differences, accents, co-articulation and speaking rate, etc. \cite{eisner}. In this study, we investigate whether a computational model of speech recognition inspired by infant learning processes can learn to recognise words without prior linguistic knowledge. 

Cognitive science has long tried to explain our capacity for speech comprehension through computational models (see \cite{weber2012models} for an overview). Models such as Trace \cite{elman1988}, Cohort \cite{marlsenwilson}, Shortlist \cite{norris1994}, Shortlist B \cite{NorrisMcqueen} and FineTracker \cite{scharenborg2010} attempt to explain how variable and continuous acoustic signals are mapped onto a discrete and limited-size mental lexicon. These models all assume that the speech signal is first mapped to a set of pre-lexical units (e.g., phones, articulatory features) and then mapped to a set of lexical units (words). However, the exact set of units is predetermined by the model developer, avoiding the issue of learning what these units are in the first place. Even the recently introduced DIANA model \cite{tenbosch}, which does away with fixed pre-lexical units, still uses a set of predetermined lexical units. While all these models have proven successful at explaining behavioural data from listening experiments, they all require prior knowledge in the form of a fully specified set of (pre-)lexical units. 

The fact that infants are able to learn words without explicit supervision suggests that it should be possible for computational models to do so in a similar manner. The model that we investigate in the current work exploits visual context in order to learn to recognise words in speech without supervision or prior knowledge of words. 

\subsection{Visually grounded speech}

Humans have access to multiple streams of sensory information besides the speech signal, perhaps most prominently the visual stream. Speech is often used to refer to and describe the world around us. It is theorised that infants learn to extract their words from speech by repeatedly hearing words while seeing the same objects or actions \cite{rasanen2015joint}. For instance, parents might say `the ball is on the table' and `there's a ball on the floor' etc., while consistently pointing towards a ball.

Visually Grounded Speech (VGS) models are speech recognition models inspired by this learning process. The basic idea behind VGS models (e.g. \cite{dedeyne2021,HarwathVQ,kamper2019semantic}) is to make use of co-occurrences in the visual and auditory streams. For instance, from the sentences `a dog playing with a stick' and `a dog running through a field' along with images of these scenes, a model could learn to link the auditory signal for `dog' to the visual representation of a dog because it is common to both image-sentence pairs. This allows the model to discover words, meaning it learns which utterance constituents are meaningful linguistic units of their own. While there is a wide variety of VGS models, they all share the common concept of combining visual and auditory information in a common multi-modal representational space in which the similarity between matching image-sentence pairs is maximised while the similarity between mismatched pairs is minimised. 

The potential of visual input for modelling the learning of linguistic units has long been recognised. In 1998, Roy and Pentland introduced their model of early word learning \cite{Roy1998}. While many models at the time (and even today) relied on phonetic transcripts or written words, they implemented a model that learns solely from co-occurrences in the visual and auditory input. Their model builds an `audio-visual lexicon' by finding clusters in the visual input and looking for reoccurring segments in the acoustic signal. Their model performs many tasks that are still the focus of research today: unsupervised discovery of linguistic units, retrieval of relevant images and generation of relevant utterances. However, the model was limited limited to colours and shapes (utterances like `this is a blue ball') and does not show it can learn from more natural less restricted input. 

The tasks performed by Roy and Pentland's involve challenges for both computer vision and natural language processing, and advances in both fields have renewed interest in multi-modal learning and with it the need for multi-modal datasets. In 2013, Hodosh, Young and Hockenmaier introduced Flickr8k \cite{Hodosh2015}, a database of images accompanied by written captions describing their content, which was quickly followed by similar databases such as MSCOCO Captions \cite{Chen2015}. These datasets are now widely used for image-caption retrieval models (e.g., \cite{merkx2019,Karpathy2017,Klein2015,Ma2015,Vendrov2015,Wehrmann2018,Dong2018}) and caption generation (e.g., \cite{Karpathy2017,Xu2015}).

Harwath and Glass collected spoken captions for the Flickr8k database and used it to train the first neural network based VGS model \cite{Harwath2015}. There have been many improvements to the model architecture (\cite{Harwath2016,Chrupala2017,merkxinterspeech,Havard2020,harwath2018jointly,scharenborg2020,kamper2018visually}) and new applications of VGS models such as semantic keyword spotting (\cite{Kamper2017a,Kamper2017b,kamper2019semantic}), image generation \cite{Wang2021}, recovering of masked speech \cite{Srinivasan2020} and even models combining speech and video \cite{Palaskar2018}. 

Many studies have since investigated the properties of the learned representations of such VGS models (e.g., \cite{HarwathVQ,Chrupala2018,Hsu2019,chrupala2020analyzing,merkx2021}). Perhaps the most prominent question is whether words are encoded in these utterance embeddings even though VGS models are not explicitly trained to encode words and are only exposed to complete sentences. The representations created by the VGS model presented in \cite{harwath2018jointly} show that speech segments are often most similar to visual patches corresponding to their visual referents. In \cite{Chrupala2017,merkxinterspeech}, the authors show that VGS models encode the presence of individual words that can reliably be detected in the resulting sentence representation.

R\"as\"anen and Khorrami \cite{Rasanen2019} made a VGS model that was able to discover words from even more naturalistic input than image captions: recordings made by head-mounted cameras worn by infants during child-parent interaction. The authors showed that their model was able to learn utterance representations from which several words (e.g., `doggy', `ball') could reliable be detected. Even though their model used visual labels indicating the objects the infants were paying attention to rather than the actual video input, this study is an important step towards showing that VGS models can acquire linguistic units from actual child-directed speech. 

While the presence of individual words is encoded in the representations of a VGS model, the model does not explicitly yield any segmentation or discrete linguistic units. A technique which allows for the unsupervised acquisition of such discrete units is Vector Quantisation (VQ). VQ layers were recently popularised by \cite{vanoord}, who showed that these layers could efficiently learn a discrete latent representational space. Harwath, Hsu and Glass \cite{HarwathVQ} have recently applied these layers in a VGS model, and showed that their model learned to encode phonemes and words in its VQ layers. 

Havard and colleagues went further than simply detecting the presence of words in sentence representations. They presented isolated nouns to a VGS model trained on whole utterances and showed that the model was able to retrieve images of their visual referents \cite{havard2019word}. This showed that their model did not just encode the presence of these constituents into the sentence representations, but actually `recognised' individual words and learned to map them onto their correct visual referents. So regarding the example mentioned above, the model learned to link the auditory signal for `dog' to the visual representation of a dog.

However, the model by Havard and colleagues \cite{havard2019word} was trained on synthetic speech. Word recognition in natural speech is known to be more challenging, as shown for instance by the large performance gap between the VGS models trained on synthetic and real speech in \cite{Chrupala2017}. Dealing with the variability of speech is an important aspect of human speech recognition. If VGS models are to be plausible as computational models of speech recognition, it is important that these models implicitly learn to extract words from natural speech. 

\subsection{Current study}

The goal of this study is to investigate whether a VGS model discovers and recognises words from natural speech without prior linguistic information. We furthermore investigate the model's cognitive plausibility by testing whether its word recognition performance is affected by word competition effects known to take place during human speech comprehension. We do so by answering the following questions: 1) does a VGS model trained on natural speech learn to recognise words, and does this generalise to isolated words, 2) is the model's word recognition affected by word competition effects, 3) does the model learn the difference between singular and plural nouns, and 4) does the introduction of VQ layers in order to learn discrete linguistic units aid word recognition?

Our first experiment is a continuation of our previous work\cite{scholten} and the work by Havard et al. \cite{havard2019word}. We train and test our VGS models on natural speech, as opposed to the synthetic speech used in \cite{havard2019word}. Furthermore whereas previous work focused on the recognition of nouns, we also include verbs as our target words. As in \cite{havard2019word}, we present isolated target words to the VGS model and use the retrieval of images to measure the model's word recognition performance by looking at the proportion of retrieved images containing a word's correct visual referent. If the model is indeed able to recognise this word in isolation, it should be able to retrieve relevant images depicting the word's visual referent, indicating that the model has learned a representation of this word based on the multi-modal input. We also investigate the influence of linguistic and acoustic factors on the model's recognition performance using generalised linear mixed effects regression. For instance, we know that faster speaking rates have a negative impact on human word recognition performance (e.g. \cite{koch2016}). For this experiment we collected new speech data, consisting of words pronounced in isolation. On the one hand, such data can be thought of as `cleaner' than words extracted from sentences (as in \cite{scholten}) due to the absence of co-articulation. On the other hand, the model has only seen words in their sentence context, co-articulation included, and might rely on this contextual information too heavily to be able to recognise words in isolation. Thus this allows us to investigate whether a VGS model learns to recognise words independently of their context, answering our first research question.

In our second experiment we investigate the time course of word recognition in our model. This allows us to test whether the word recognition performance of our VGS model is affected by word competition as is known to take place during human speech comprehension. For this experiment, we look at two measures of word competition, namely, word-initial cohort size and neighbourhood density. In the Cohort model of human speech recognition, the incoming speech signal is mapped onto phone representations. These activated phone representations activate every word in which they appear and, as more speech information becomes available, activation reduces for words that no longer match the input. The word that best matches the speech input is recognised. The number of activated or competing words is called the word-initial cohort size and plays a role in human speech processing: the more competitors there are, the longer it takes to recognise a word \cite{norris1995competition}. Words with a denser neighbourhood of similarly sounding words are also harder to recognise as they compete with more words \cite{Luce1998}. 

We also use our model to test the interaction between neighbourhood density and word count. There have been several studies that investigated this interaction with inconclusive results. In a gating study, Metsala \cite{metsala} found an interaction where for low-frequency words, recognition was facilitated by a dense neighbourhood and recognition of high-frequency words was facilitated by a sparse neighbourhood. Goh et al. found that response latencies in word recognition were shorter for words with sparse neighbourhoods than for words with dense neighbourhoods \cite{goh}. They furthermore found a higher recognition accuracy for sparse-neighbourhood high-frequency words as opposed to the other conditions (i.e., sparse-low, dense-high, dense-low). This means that, unlike Metsala, they found no facilitatory effect of neighbourhood density for low-frequency words. Furthermore, Rispens, Baker and Duinmeijer \cite{rispens} and Garlock, Walley and Metsala \cite{garlock} found no interaction between lexical frequency and neighbourhood density at all. 

For this experiment we use a gating paradigm, a well known technique borrowed from human speech processing research (e.g., \cite{cotton1984gating,Smith2017}). In the gating experiment, a word is presented to the VGS model in speech segments of increasing duration, that is, with an increasing number of phones, and the model is asked to retrieve an image of the correct visual referent on the basis of the speech signal available so far. We then use generalised linear mixed effects regression to predict word recognition performance from the word competition effects of interest and several control features.

In our third experiment we investigate whether our VGS model learns to differentiate between singular and plural instances of nouns. By the same principle of co-occurrences in the visual and auditory streams that allows the model to discover and recognise words, it may also be able to differentiate between their singular and plural forms. We test this by presenting both forms of all nouns to the model, and analysing whether the retrieved images contain single or multiple visual referents of that noun.

Our fourth question is aimed at investigating VQ, a technique that was recently first applied to VGS models by Harwath, Hsu and Glass \cite{HarwathVQ}. While discrete linguistic units (including words) were indeed acquired by their model, it is unclear if this knowledge generalises to recognising words in isolation. If so, we expect that the addition of VQ layers improves word recognition results of our VGS model. Havard, Chevrot and Besacier \cite{Havard2020} for instance, provided explicit word boundary information to their VGS model which improved its performance, showing that knowledge of the linguistic units is beneficial to the model. Rather than explicitly providing this word boundary information as in \cite{Havard2020}, VQ layers allow segments to emerge in an end-to-end fashion without providing additional prior knowledge. This makes VQ layers a more suitable approach for our model as prior knowledge of word boundaries is not cognitively plausible. In order to investigate if the introduction of VQ layers aids word recognition we do not conduct a separate experiment, but compare our baseline VGS model to a VGS model with added VQ layers throughout the other experiments. 

\section{Methods}

\subsection{Visually Grounded Speech model}

\subsubsection{Model architecture}

\begin{figure*}
    \centering
    \includegraphics[width =\linewidth]{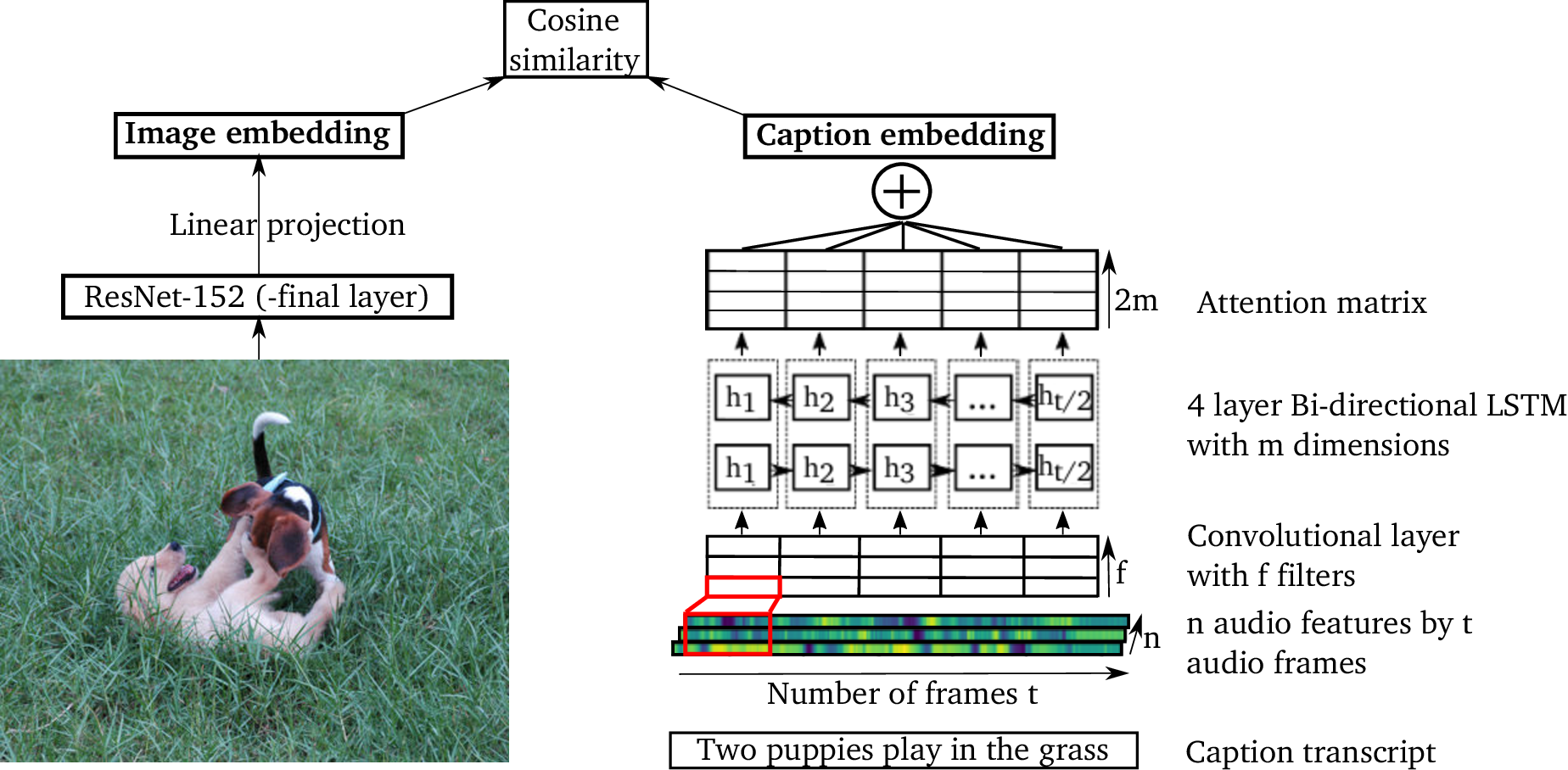}
    \caption{Model architecture: the model consists of two branches with the image encoder depicted on the left and the caption encoder on the right. The audio features consist of 13 MFCC with 1st and 2nd order derivatives by $t$ frames. Each LSTM hidden state $\mathbf{h}_{t}$ has $1024$ features which are concatenated for the forward and backward LSTM into $2048$ dimensional hidden states. Vectorial attention weighs and sums the hidden states resulting in the caption embedding. The linear projection in the image branch maps the image features to the same 2048 dimensional space as the caption embedding. Finally, we calculate the cosine similarity between the image and caption embedding.}
    \label{network}
\end{figure*}

Our VGS model consists of two deep neural networks as depicted in Figure \ref{network}; one to encode the images and one to encode the audio captions. The model is trained to embed both input streams in a common embedding space with the goal of minimising the cosine distance between image-caption pairs in this space while at the same time maximising the distance between mismatched pairs. We do not further fine-tune the hyper-parameters of the model and use the best parameters found in \cite{merkx2019} because the goal of this study is not to improve the training task score. Our goal is to perform experiments in order to learn more about the unsupervised discovery and recognition of linguistic units in such a model. 

It is common practice to use a pre-trained image recognition network for the image branch of a VGS model (e.g., \cite{Kamper2017b,Chrupala2017,HarwathVQ}). We use the ResNet-152 network \cite{He2016}, which is a pre-trained convolutional network that was trained on ImageNet \cite{imagenet_cvpr09}, to extract our image features. We take the activations of the penultimate fully connected layer by removing the final object classification layer from this network and apply a single linear layer of size 2048 to these outputs. Finally, we normalise the results to have unit L2 norm. The goal of the linear projection is to map the ResNet-152 features to the same 2048-dimensional embedding space as the audio representations. The image embedding $\mathbf{i}$ is given by:
\begin{equation}
\mathbf{i} = \frac{\mathbf{img}A^T+\mathbf{b}}{||\mathbf{img}A^T+\mathbf{b}||_2}
\end{equation}
where $A$ and $\mathbf{b}$ are learned weight and bias terms, and $\mathbf{img}$ is the vector of ResNet-152 image features. 

The audio branch consists of a 1-d convolutional neural network of size 6, stride 2 and 64 output channels, which sub-samples the signal along the temporal dimension. The resulting features are fed into a 4-layer bi-directional LSTM with 1024 units.\footnote{In \cite{merkxinterspeech} we used a 3-layer GRU, but it has since then become practically feasible to train larger models on our hardware.} The 1024 bi-directional units are concatenated to create a 2048 feature vector. The self-attention layer computes a weighted sum over all the hidden LSTM states: 
\begin{equation}
    \mathbf{a}_t = \softmax(V\tanh(W\mathbf{h}_t +\mathbf{b}_w)+\mathbf{b}_v) 
    \label{eq3}
\end{equation}

where $\mathbf{a}_t$ is the attention vector for hidden state $\mathbf{h}_t$, and $W$, $V$, $\mathbf{b}_w$, and $\mathbf{b}_v$ indicate the weights and biases. The learnable weights and biases are implemented as fully connected linear layers with output sizes 128 and 2048, respectively. The applied attention is then the sum over the Hadamard product between all hidden states $(\mathbf{h}_1, ..., \mathbf{h}_t)$ and their attention vector:
\begin{equation}
    \Att(\mathbf{h}_1, ..., \mathbf{h}_t) = \sum\limits_{t}\mathbf{a}_t\circ\mathbf{h}_t
   \label{eq4}
\end{equation}

The resulting embeddings are normalised to have unit L2 norm. The caption embedding $\mathbf{c}$ is thus given by:
\begin{equation}
\mathbf{c} = \frac{\Att(\LSTM(\CNN(\mathbf{a}_1, ..., \mathbf{a}_t)))}{||\Att(\LSTM(\CNN(\mathbf{a}_1, ..., \mathbf{a}_t)))||_2} 
\end{equation}

where $\mathbf{a}_1, ..., \mathbf{a}_t$ indicates the caption represented as $t$ frames of MFCC vectors and $\Att$, $\LSTM$ and $\CNN$ are the attention layer, stacked LSTM layers, and convolutional layer, respectively. 

% at end of doc for review

Next, we also implement a VGS model with added VQ layers \cite{vanoord}. We will refer to our regular model and the model with VQ layers as LSTM and LSTM-VQ models, respectively. Our implementation most closely follows \cite{HarwathVQ}, who were the first to apply these layers in a VGS model, and showed that their model learned discrete linguistic units. VQ layers consist of a `codebook' which is a set of $n$-dimensional embeddings. A VQ layer discretises incoming input by mapping it to the closest embedding in the codebook and passing this embedding to the next layer:
\begin{equation}
    VQ(\mathbf{x}) = \mathbf{e}_k, \text{  where   } k = \text{argmin}_j || \mathbf{x} - \mathbf{e}_j ||_2
\end{equation}
with $\mathbf{x}$ being the VQ layer input and $\mathbf{e}_j$ being the codebook embeddings. 

For the LSTM-VQ model we insert VQ layers in the LSTM stack after the first and after the second LSTM layer. We use two layers as \cite{HarwathVQ} showed that when using two VQ layers, a hierarchy of linguistic units emerges: the first layer best captures phonetic identity while in the second layer, several codes emerged which are sensitive to specific words. The first and second VQ layers have 128 and 2048 codes respectively.  

We use our own PyTorch implementation of the models and the VQ layer described here, adapted from our previous work presented in \cite{merkx2019,merkxinterspeech}, which is in turn most closely related to, and based on, the VGS models presented in \cite{Harwath2016,Chrupala2017}.
Our implementation can be found on \url{removed for review}.

\subsubsection{Training data}

We train the model on Flickr8k, a well-known dataset of images with spoken captions \cite{Hodosh2015}. This is a database of 8,000 images from the online photo sharing platform Flickr.com for which five English written captions are available. Annotators were  asked  to  ‘write  sentences  that  describe  the  depicted scenes,  situations,  events  and  entities  (people,  animals,  other  objects)’ \cite{Hodosh2015}. We will use the spoken captions provided by \cite{Harwath2015}, who collected these spoken captions by having Amazon Mechanical Turk (AMT) workers pronounce the original written captions. We use the data split provided by \cite{Karpathy2017}, with 6,000 images for training and a development and test set both of 1,000 images. 

Image features are extracted by resizing all images while maintaining the aspect ratio such that the smallest side is 256 pixels. Ten crops of 224 by 224 pixels are taken, one from each of the corners, one from the middle and similarly for the mirrored image. We use ResNet-152 \cite{He2016} to extract visual features from these ten crops and then average the features of the ten crops into a single vector with 2,048 features.

The audio input consists of Mel Frequency Cepstral Coefficients (MFCCs). We compute the MFCCs using 25 ms analysis windows with a 10 ms shift. The MFCCs were created using 40 Mel-spaced filterbanks. We use 12 MFCCs and the log energy feature and add the first and second derivatives resulting in 39-dimensional feature vectors. Lastly we apply per utterance cepstral mean and variance normalisation.

\subsubsection{Training}

The model is trained to embed the images and captions such that the cosine similarity between image and caption embeddings is larger for correct pairs than the similarity between mismatching pairs. This batch hinge loss $L$ as a function of the network parameters \(\theta\) is given by:
\begin{equation}
    \begin{aligned}L(\theta) = \sum\limits_{(\mathbf{c},\mathbf{i}),(\mathbf{c}',\mathbf{i}')\in B} \biggl(\max(0, \cos(\mathbf{c},\mathbf{i}') - \cos(\mathbf{c},\mathbf{i}) + \alpha) +\\
    \max(0, \cos(\mathbf{i},\mathbf{c}') - \cos(\mathbf{i},\mathbf{c}) + \alpha)\biggr) \end{aligned}
\end{equation}

where $(\mathbf{c},\mathbf{i})\neq(\mathbf{c}',\mathbf{i}')$, $B$ is a minibatch of correct caption-image pairs $(\mathbf{c},\mathbf{i})$, where the other caption-image pairs in the batch serve to create mismatched pairs $(\mathbf{c},\mathbf{i}')$ and $(\mathbf{c}',\mathbf{i})$. We take the cosine similarity and subtract the similarity of the mismatched pairs from the matching pairs such that the loss is only zero when the matching pair is more similar than the mismatched pairs by a margin $\alpha$, which was set to $0.2$. 

Training task performance is evaluated by caption-to-image and image-to-caption retrieval Recall@N. For these retrieval tasks the caption embeddings are ranked by cosine distance to the image and vice versa where Recall@N is the percentage of test items for which the correct image or caption was in the top N results. Furthermore we evaluate the median rank of the correct image or caption.  

Because the VQ operation is indifferentiable, a trick called \emph{straight through estimation} is required to pass a learning signal to layers before the VQ layer \cite{bengio2013}. Put simply, as there is no gradient for the VQ operation, the gradients for the VQ output are copied and used as an approximation of the gradients for the VQ input. 

The VQ layer learns to make the codes in the codebook more similar to its inputs and vice versa. The first is accomplished by an exponential moving average. When an embedding is activated, it gets multiplied by a decay factor $\gamma$ and summed with $(1-\gamma)\mathbf{x}$ where $\mathbf{x}$ is the input that activated the embedding. Making the inputs more similar to the embeddings is accomplished by a separate VQ loss, which is the mean squared error between each input and its closest embedding.

The networks are trained using Adam \cite{Kingma2015} with a cyclic learning rate schedule based on \cite{Smith2017}. The learning rate schedule varies the learning rate smoothly between a minimum and maximum bound, which were set to $10^{-6}$ and $2\times10^{-4}$, respectively. 

We train the regular LSTM-based network for 16 epochs. Following \cite{HarwathVQ}, we \emph{warm start} the LSTM-VQ model by taking the trained LSTM network, inserting the VQ layers and training for another 16 epochs. While, unlike \cite{HarwathVQ}, we did not encounter a large performance loss for \emph{cold started} networks, we did find that a cold started VQ network frequently suffered from codebook collapse \cite{Niekerk2020}. This is an issue where suddenly all VQ inputs are mapped to only a few (often even just one) codes and from which the model never recovers.

We trained 20 VGS models of each type (with and without VQ) using different seeds for the pseudo-random number generator, in order to account for random effects of the weight initialisation and order in which the training data is presented.

\subsection{Data collection}

\subsubsection{Target words}

Visually grounded speech models exploit the fact that words in the speech signal tend to co-occur with visual referents in their corresponding images. We can therefore expect that if the system indeed learns to recognise words, they will be words with clear visual referents in the images, so we limit our analysis to the recognition of nouns and verbs. We only look at high-frequency words that the model has had ample opportunity to learn to recognise. 

We selected the 50 nouns and 50 verbs with the most frequent lemma in the Flickr8k database, excluding some words like `air' and `stand' as they appear in nearly every picture and as such recognition cannot be properly measured. Other examples of rejected words are verbs such as `try' since it is not possible to set clear objective standards for the visual referent of this verb. The selected words are shown in Table \ref{words}

% at end of doc for review

\begin{table}
    \caption{Selected target nouns and verbs in order of occurrence in the training set transcripts. A * indicates nouns for which only a single form was recorded, + indicates words that were not included in the analysis because there were not enough images depicting their visual referent in the test set.}
    \begin{tabular}{l l l l | l l l l}
        \hline
        \multicolumn{4}{l}{Nouns} &\multicolumn{4}{l}{Verbs}\\ \hline
        dog	& & man & & play & & run & \\ 
        boy	& & girl & & jump & & sit &\\
        woman & & water* & & hold & & walk &\\
        shirt & & ball & & ride & & climb &\\ 
        grass* & & beach & & smile & & pose &\\
        snow* & & group & & catch & & carry &\\
        street & & rock & & leap & & perform &\\
        camera & & bike & & fly & & dance &\\
        mountain & & hat & & swim & & eat &\\
        pool & & player & & pull & & hang &\\
        jacket & & ocean & & chase & & slide &\\
        basketball & & sand* & & splash & & point &\\
        car & & building & & kick & & throw &\\
        soccer* & & swing & & fight & & swing &\\
        football & & sunglasses* & & lie & & lay &\\
        shorts* & & park & & laugh & & ski &\\ 
        dress & & table & & surf & & drive &\\
        hand & & tree & & fall & & follow &\\
        lake & & hill & & race & & roll &\\
        toy & & baby & & hit & & reach &\\ 
        tennis*+ & & river & & wade & & lean &\\
        wave & & snowboarder & & push & & bite &\\
        bench & & game & & spray & & paddle &\\
        surfer & & stick & & light+ & & bend &\\
        team & & skateboard & & cross & & raise &\\
        \hline
    \end{tabular}
    \label{words}
\end{table}

To test word recognition performance, we present the selected target verbs and nouns in isolation. Two North American native speakers of English (one male, one female) were asked to read the target words out loud from paper. The words were recorded in isolation by asking the speakers to leave at least a second of silence in between words. Both speakers are not present in the Flickr8k database. In order to keep conditions close to those of the Flickr8k spoken captions (and other captioning databases collected through AMT), the speakers recorded the sentences at home using their own hardware. They were asked to find a quiet setting and record the words in a single session. 

The nouns were presented in both their singular and plural form (where applicable)\footnote{`Shorts' and `sunglasses' are syntactically plural, but we group them under the singular nouns as their use in the data is most often in reference to a single object.}.
All verbs were recorded in root form, third person singular form, and progressive participle. We did not record past tense verb forms as these are rarely, if ever, used in the image descriptions. The speakers received a \$20 gift card for their participation. 

The speech data were recorded in stereo at 44.1kHz in Audacity. We down-sampled the utterances to 16kHz and converted them to mono to match the conditions of the Flickr8k captions, after which we applied the same MFCC processing pipeline used for the Flickr8k training data.

\subsubsection{Image annotations}\label{img_annotations}

We test whether the VGS models learned to recognise the recorded target words by presenting them to the model and checking whether the retrieved images contain the visual referents of the target words. The problem with this approach, however, is that Flickr8k contains no ground truth image annotations that might be used for such a test. The captions might serve as an indication: if annotators mention an action or object in the caption we can be reasonably sure it is visible in the picture. In contrast, it is definitely not the case that if an object or action is not mentioned, it is not in the picture, leading to an underestimation of model performance.

We created a ground truth labelling for the visual referents of our target words by manually annotating the 1000 images in the Flickr8k test split. For the nouns, we also indicate whether the visual referent occurred only once or multiple times in the images, allowing us to test whether the model properly learns to differentiate between the plural and singular forms of the noun. 

Annotations were made by two annotators, one covering the nouns and one the verbs. In order to check the quality of the annotations, the fist author annotated a sample of $5\%$ of the images. We calculate an inter annotator agreement based on this sample (Verbs: $\kappa=0.70$, Nouns: $\kappa=0.76$). 

\subsection{Word recognition}\label{p@n}

Following \cite{havard2019word}, we use the retrieval of images containing a target word's correct visual referent as a measure of the model's word recognition performance. As this is a retrieval task where multiple correct images can be found per word, we use precision@10 (P@10) to measure word recognition performance. That is, for each target word embedding we calculate the cosine similarity to all test image embeddings and retrieve the ten most similar images. Precision@10 is then the percentage of those images that contains the correct visual referent according to our annotations. We excluded two of our target words from this analysis as there were fewer than ten test images containing their visual referent. Although we annotated whether an image contains a single or multiple visual referents, unless specifically stated, multiple visual referents were counted as correct for a singular noun and vice versa for the purpose of calculating P@10. 

We also create P@10 scores for two baseline models. Our random baseline is simply the averaged score over five randomly initialised and untrained VGS models. This results in a random selection of images but since some words' visual referents occur in dozens to hundreds of test images, the recognition scores are far from zero. Our naive baseline represents the best recognition score a model could possibly get if it always retrieved the ten images with the highest number of visual referents (i.e., always the same ten images, selected separately for the nouns and verbs). Note that this baseline is not realistic and requires knowledge of the contents of the test set (namely the number of visual referents per image). Still, it is useful to compare our model performance to a model that has only a single response regardless of the input query. 

We then examine the influence of linguistic and acoustic factors on the model's word recognition performance as measured by P@10, using a Generalised Linear Mixed Model (GLMM) with beta-binomial distribution\footnote{Our P@10 data, which is discrete and has a floor of 0 and a ceiling of 10, is not suited for standard linear modelling. Our response variable is best described as a series of Bernoulli trails with successes and failures in terms of correct and incorrect retrieval.} and canonical logit link function. We used the \textit{glmmTMB} package in R \cite{glmmTMB}.

The GLMM examines the effects of signal duration (i.e., number of speech frames), speaking rate (number of phonemes per second), number of vowels, number of consonants, morphology (singular or plural)\footnote{As seen in Section \ref{word_recog}, word recognition results on the verbs were overall a lot worse than for the nouns and we decided not to continue our analysis on the verbs.} and VQ (LSTM or LSTM-VQ model) with the VGS model's word recognition performance (P@10) as the outcome variable. We furthermore include the (log-transformed) counts of the target word and its lemma in the training set as we expect better recognition for words that are seen more often during training. The correlation between lemma count and word count is .48, so they are expected to explain unique portions of variance. We also include speaker-ID to account for differences in recognition performance between the two speakers. Number of vowels and consonants are centered, all other non-categorical variables are standardised. VQ and morphology were dummy coded and speaker ID was effect coded. 

The GLMM included by-lemma and by-model (each of the 20 random initialisations) random intercepts. We started by including all fixed effects that vary within lemma and model-ID as by-lemma and by-model random slopes but this model was unable to converge. As a maximal model is thus not possible, we used the following procedure to reduce the model complexity until the model converged.

We started by using a zero-correlation-parameter GLMM, which still did not converge. Next, we tried two separate GLMMs, one with all uncorrelated by-lemma random slopes and one with all uncorrelated by-model random slopes. The by-model GLMM results in a singular fit for the speaker ID, morphology, and VQ random slopes. After removing these by-model slopes, the combined GLMM with all remaining uncorrelated by-lemma and by-model slopes converged. None of the removed random slopes could be added back into the combined GLMM without causing convergence issues. The final GLMM formula is: 

\texttt{
p@10 $\sim$ speaking rate + duration + \\lemma count + word count + \#vowels + \\\#consonants + VQ + speaker id + morphology + \\(1 + speaking rate + duration + word count + \\\#vowels + \#consonants + VQ + speaker id + \\morphology || lemma) + (1 + speaking rate + \\duration + lemma count + word count + \#vowels \\+ \#consonants || model id)
}

\subsection{Word competition}

We perform a gating experiment to investigate word competition in our models. We present the models with the target words in segments of increasing length, using one gate per phoneme. Simply put, if the target word is `dog' with the phonemes /d-\textipa{O}-g/, we evaluate performance after the model has processed /d/, /d-\textipa{O}/, and finally the whole word /d-\textipa{O}-g/. Performance is measured in P@10 as described in \ref{p@n}. 

For the gating experiment we need to know when each phoneme starts and ends. We use the Kaldi toolkit to make a forced alignment of our target words and their phonetic transcripts \cite{Povey2011}. The phonetic transcripts of our target words are taken from the CMU Pronouncing Dictionary available at \url{www.speech.cs.cmu.edu/cgi-bin/pronounce}. 

We test the competition effects by defining the word-initial cohort of a target word as the words in the Flickr8k dataset that share the same word-initial phoneme sequence as those seen so far at the current gate. That is, the number of words in the word-initial cohort equals the number of words that cannot be distinguished from one another given the sequence so far, and thus the number of words competing for recognition. 

We define neighbourhood density as the number of words that differ exactly one phoneme from the target word \cite{vitevitch2016phonological}. These words are expected to compete for recognition, and the number of words that strongly resemble the target word influences word recognition. Research shows that words with a dense neighbourhood are harder to recognise than those with a sparse neighbourhood \cite{Luce1998}. 

For both the word-initial cohort and the neighbourhood density, we use phonetic transcripts from the CMU pronouncing dictionary, which contains the transcripts for a total of 6431 words in the Flickr8k captions. 

We test whether the neighbourhood density and word-initial cohort size affect word recognition in our model using a GLMM. Furthermore, we are interested in three interaction effects. As previously discussed, we want to test the interactions between neighbourhood density and the word and lemma counts. As for the third interaction, we are interested in the interaction between VQ and the number of phonemes processed so far (gate number). The VGS model with VQ layers is forced to map its inputs to discrete units even as early as the first gate. As the second VQ layer has been shown to learn discrete word-like representations \cite{HarwathVQ}, we might expect that words are recognised earlier, as would be indicated by a smaller effect of gate number for the LSTM-VQ model. 

The GLMM's fixed effects are the neighbourhood density, gate number, the size of the word-initial cohort, VQ, morphology, the number of vowels and the number of consonants. Again we also add the frequencies of occurrence of the target word and its lemma in the training set and subject-ID to account for expected effects of training data frequency and speaker differences. The number of vowels, number of consonants and gate number are centered, all other non-categorical variables are standardised.

The GLMM has by-lemma and by-model random intercepts. We started with maximal by-lemma and by-model random slopes but had to reduce the complexity in due to convergence issues. As before, we started with uncorrelated random slopes but this model failed to converge. We then fit two separate GLMMs, one with all uncorrelated by-lemma random slopes and one with all by-model random slopes. The by-model GLMM resulted in a singular fit for the speaker-ID, morphology, and VQ random slopes, which had to be removed. After the removal of these random slopes the combined model still failed to converge. We proceeded to use the variance estimates of the separate GLMMs to remove the smallest variance components until the combined GLMM was able to converge. This led to the removal of all by-model random slopes and the by-lemma slopes for number of vowels and word count. The final GLMM formula for analysis of the gating experiment is:

\texttt{
p@10 $\sim$ (lemma count + word count) *\\ density + VQ * gate + initial cohort size + \\ speaker id + morphology + \#vowels + \\ \#consonants + (1 + density + VQ + gate + \\ initial cohort size + speaker id + morphology \\ + \#consonants || lemma) + (1 | model id)
}

\section{Results}

All results presented here are averaged over the 20 random initialisations of the VGS model. We first evaluate how well the models perform on the training task and compare their performance to other VGS models. The scores in Table \ref{flickr_c2i_results} show the result for the speech caption-to-image and image-to-caption retrieval tasks. This indicates how well the model learned to embed the speech and images in the common embedding space. As expected, the VQ layers are beneficial to the VGS model's training task performance \cite{HarwathVQ}. 

% at end of doc for review

\begin{table}
    \caption{Image-caption retrieval results on the Flickr8k test set. R@N is the percentage of items for which the correct image or caption was retrieved in the top N (higher is better) with 95\% confidence interval. Med r is the median rank of the correct image or caption (lower is better). We compare our VGS models to previously published results on Flickr8k. `-' means the score is not reported in the cited work.}
    \centering
    \resizebox{\linewidth}{!}{
        \begin{tabular}{l | r r r r }

            \multicolumn{1}{l}{Model} & \multicolumn{4}{c}{Caption to Image} \\ \hline

             & R@1 & R@5 & R@10 & med r \\ \hline
            \cite{Harwath2015} & - & - & 17.9$\pm$1.1& - \\
            \cite{Chrupala2017} & 5.5$\pm$0.6 & 16.3$\pm$1.0 & 25.3$\pm$1.2 & 48 \\ 
            \cite{merkxinterspeech} & 8.4$\pm$0.8 & 25.7$\pm$1.2 & 37.6$\pm$1.3 & 21 \\ 
            \cite{Wang2021} & 10.1$\pm$0.8 & 28.8$\pm$1.3 & 40.7$\pm$1.4 & -\\
            LSTM & 12.5$\pm$0.2 & 33.8$\pm$0.3 & 46.8$\pm$0.3 & 12 \\ 
            LSTM-VQ & 12.9$\pm$0.2 & 34.5$\pm$0.3 & 47.3$\pm$0.3 & 12 \\ \hline
            \multicolumn{1}{l}{Model} & \multicolumn{4}{c}{Image to Caption} \\ \hline
             & R@1 & R@5 & R@10 & med r \\ \hline
            \cite{Harwath2015} & - & - & 24.3$\pm$2.7 & - \\
            \cite{merkxinterspeech} & 12.2$\pm$2.0 & 31.9$\pm$2.9 &
            45.2$\pm$3.1 & 13 \\ 
            \cite{Wang2021} & 13.7$\pm$2.1 & 36.1$\pm$3.0 & 49.3$\pm$3.1 & -\\
            LSTM & 18.5$\pm$0.5 & 42.4$\pm$0.7 & 55.8$\pm$0.7 & 8 \\ 
            LSTM-VQ & 19.6$\pm$0.6 & 45.4$\pm$0.7 & 58.1$\pm$0.7 & 7 \\ 
            \hline

        \end{tabular}
    \label{flickr_c2i_results}
    }
\end{table}

\subsection{Word recognition}\label{word_recog}

In the first experiment, we presented isolated words to the model. Table \ref{recog_results} shows the average P@10 scores. The singular nouns are recognised best with P@10 scores of .519 and .529 for the LSTM and LSTM-VQ model, respectively. This means that on average more than five out of the ten retrieved images contain the correct visual referent. For the plural nouns the average performance is .479 and .449 for the LSTM and LSTM-VQ model, respectively. However, seven target nouns have no plural form, so the scores for plural and singular nouns are not directly comparable. Therefore, we also calculate singular noun performance only on those words that also have a plural form. The results show that for the LSTM model, singular and plural forms are recognised equally well. However, it seems that the LSTM-VQ model recognises plural target words slightly less well than singular words.

\begin{figure*}
    \centering
    \includegraphics[width =\linewidth]{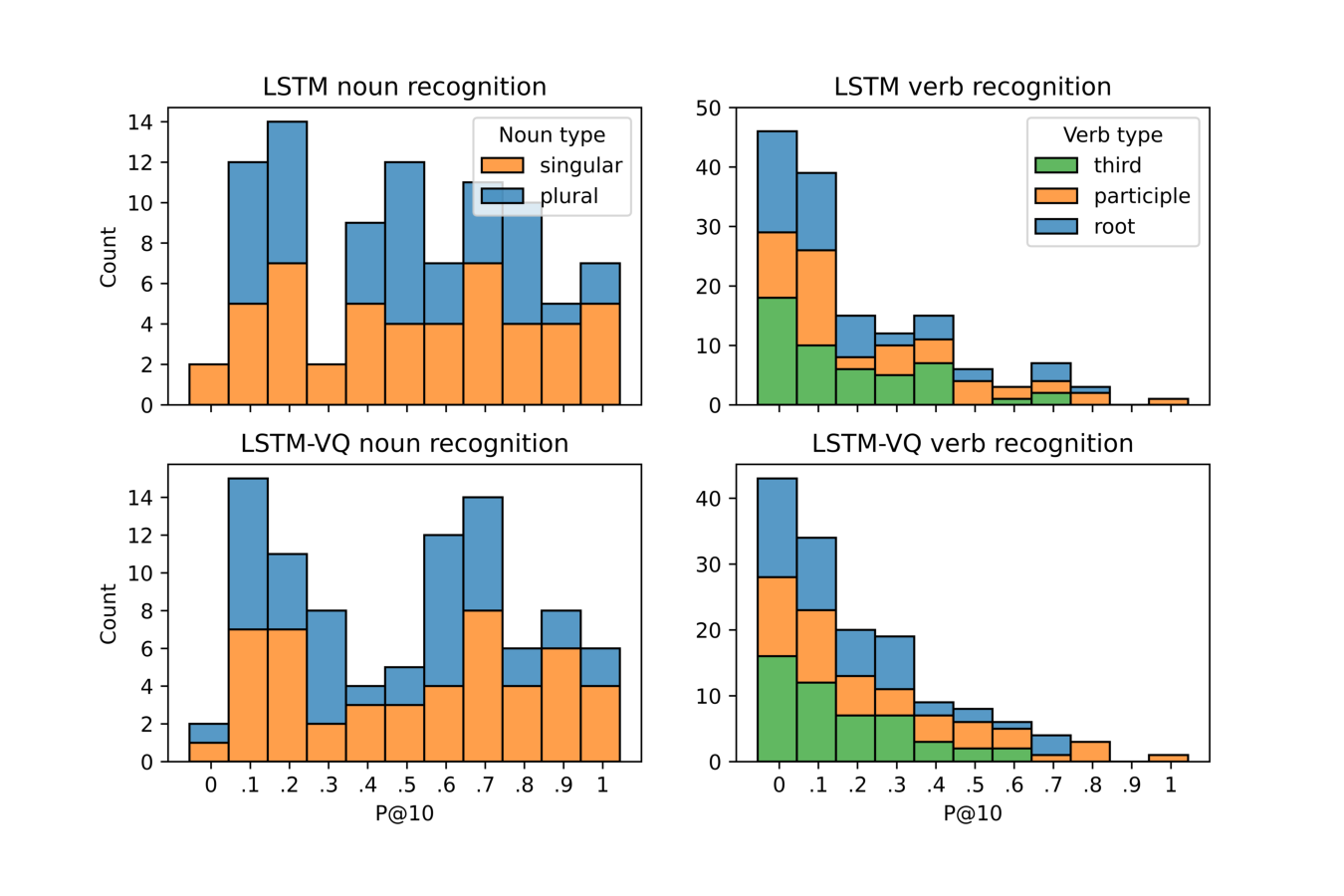}
    \caption{Histograms of the word recognition experiment results for each word type. P@10 is the percentage of items in the top ten retrieved images containing the correct visual referent for the target word, averaged over the two speakers and the random VGS model initialisations.}
    \label{recog_hist}
\end{figure*}

The histograms in Figure \ref{recog_hist} show the distribution of the P@10 scores by word type (noun or verb), morphology and whether the VGS model included VQ layers. This highlights that the recognition of the verbs is overall much worse than for the nouns: many verbs have a P@10 of zero, meaning they are not recognised at all. For the nouns on the other hand, only two words are not recognised at all. While both LSTM models recognise verbs better than the random baseline, only the participles have better performance than the naive baseline and a p@10 over .7 on some words. As the recognition performance for the verbs is obviously a lot worse than for nouns, we continued our analysis on the nouns only.

Havard and colleagues \cite{havard2019word} reported a median P@10 of 0.8 on 80 nouns (from the synthetic speech database MSCOCO), while our models achieve median p@10 scores of 0.6 and 0.5 on singular and plural nouns, respectively. Even though the models learn to recognise most nouns and even their plural forms (with only two words per model not being recognised at all), this indicates a large difference in recognition performance going from the synthetic speech dataset in \cite{havard2019word} to our real speech. Note, however, that as Havard et al. used the most frequent nouns for their dataset (MSCOCO), the target words do not fully overlap with ours. 

% at end of doc for review

\begin{table}
    \caption{Word recognition results for each noun and verb type for the trained models, a random model and a naive baseline. P@10 is the percentage of images in the top ten retrieved images that contained the correct visual referent. Between brackets are the recognition scores when only evaluating the subset of target words that also have plural forms.}
    \centering
    \resizebox{\linewidth}{!}{
        \begin{tabular}{l l l l l}
            & & & \multicolumn{2}{c}{Baseline} \\
            \multicolumn{1}{l}{Morphology} & \multicolumn{1}{c}{LSTM} & \multicolumn{1}{c}{LSTM-VQ} & Random & Naive\\ \hline
            singular noun& .519(.479)& .529(.485) & .137& .278\\
            plural noun&  .479&  .449 & .140& .267\\
            root verb&  .185  & .193 & .082& .188\\
            third-person verb& .176 &  .164 & .078 & .188\\
            participle verb& .246 &  .260 & .083 & .188\\ \hline

        \end{tabular}
    \label{recog_results}
    }
\end{table}

The results of the GLMM for the word recognition experiment are summarised in Table \ref{recog_glmm}. Speaking rate and number of consonants have a significant effect on the VGS model's word recognition performance. The positive coefficient of the number of consonants indicates that words with more consonants are on average recognised better. The negative coefficient for speaking rate indicates that words are harder to recognise if they are spoken faster. Unsurprisingly, lemma count also has a significant effect on word recognition: Lemmas that were seen more often during training are recognised better. The results further confirm that plural and singular nouns are recognised equally well and that there is no difference in recognition performance between the two speakers.

% at end of doc for review

\begin{table}
    \caption{Estimated model effects for the word recognition GLMM and the results of Type III Wald $\chi^2$ tests.}
    \centering
        \begin{tabular}{l r r r r l}
        \hline
        Effect & Estimate & Std. error & $\chi^2$ & p \\ \hline
        Intercept & $-$0.89 & 0.82 & 1.20 & 0.27 \\
        Speaking rate & $-$2.03 & 0.91 & 4.98 & \bf{0.03} \\
        Duration & $-$0.88 & 0.60 & 2.14 & 0.14\\
        Lemma count & 1.98 & 0.70 & 7.97 & \bf{0.005}\\
        Word count & 0.33 & 0.40 & 0.69 & 0.41\\
        \#Vowels & 1.33 & 1.35 & 0.98 & 0.32\\
        \#Consonants & 2.06 & 0.81 & 6.46 & \bf{0.01}\\
        VQ & -0.04 & 0.07 & 0.34 & 0.56\\
        Speaker id & 0.74 & 0.51 & 2.13 & 0.14\\
        Morphology & 0.57 & 0.88 & 0.42 & 0.52\\\hline
        \end{tabular}
    \label{recog_glmm}
\end{table}

While overall these results show no difference in word recognition performance between the LSTM-VQ and the LSTM, it is notable that the LSTM-VQ has a performance difference between singular and plural nouns, whereas the LSTM model does not. Similarly, the LSTM-VQ performs best on the participle verb form and worse on the third person and root forms. As both third person and root verbs and plural nouns are less frequent than the participle verbs and singular nouns, it may be the case that the codebook simply learns to encode frequent words better, and struggles with the less frequent word(form)s. 

To further investigate whether the VQ models were indeed better at recognising frequent words, we performed a post-hoc test where we refit the word recognition GLMM with an interaction between VQ and word count and between VQ and morphology. We fit separate GLMMs on the noun and verb data, the results of which can be seen in Table \ref{VQ_interaction}. We find the expected negative interaction between VQ and morphology where recognition on the less frequent word forms (plural, third and root) is worse than on the most frequent forms (singular, participle) for the VQ network. However, we also find an unexpected negative interaction between word count and VQ. Perhaps this is due to the correlation between word count and morphology and so partially cancels the negative effect of morphology.

% at end of doc for review

\begin{table}
    \caption{Estimated model effects for our post-hoc testing of interaction effects and the results of Type III Wald $\chi^2$ tests.}
    \resizebox{\linewidth}{!}{
    \centering
        \begin{tabular}{l r r r r l}
        \hline
        Effect & Estimate & Std. error & $\chi^2$ & p \\ \hline
        \multicolumn{5}{c}{Nouns} \\
        VQ & 0.18 & 0.05 & 16.15 & $\mathbf{<0.001}$\\
        Word count:VQ & $-$0.19& 0.04 & 23.43 & $\mathbf{<0.001}$\\
        Morphology\\\cline{1-1}
        Plural & 2.90 & 0.88 & 2.36 & 0.12\\
        Plural:VQ & $-$0.47& 0.08 & 38.76 & $\mathbf{<0.001}$\\\hline
        \multicolumn{5}{c}{Verbs} \\
        VQ & 0.22 & 0.4 & 30.88 & $\mathbf{<0.001}$\\
        Word count:VQ & $-$0.13& 0.02 & 38.42 & $\mathbf{<0.001}$\\
        Morphology\\\cline{1-1}
        Third& $-$0.22 & 0.21 \\
        Root & 0.58 & 0.53 & 4.10 & 0.13\\
        Third:VQ & $-$0.292& 0.053 \\
        Root:VQ & $-$0.141& 0.054 & 30.86 & $\mathbf{<0.001}$\\\hline
        \end{tabular}
    }
    \label{VQ_interaction}
\end{table}

\subsection{Word competition}

The results of the GLMM for the word competition experiment are summarised in Table \ref{gating_glmm}. Of the fixed effects of interest, the neighbourhood density, gate number, size of the word-initial cohort and the number of consonants have significant effects on word recognition performance. Furthermore, we found significant interaction effects between word count and neighbourhood density, and between VQ and gate number.  

As in the previous GLMM analysis the number of consonants has a positive effect. The gate number (number of phonemes processed so far) also has a positive effect. Unsurprisingly, the model is better able to recognise the target word as more of the word is seen. While this effect is modulated by the presence of VQ layers, the effect is positive for both the LSTM and the LSTM-VQ models. There is a significant negative effect of word-initial cohort size. This means recognition performance is lower the more possible candidates there are. While neighbourhood density has an overall positive effect on word recognition, care should be taken in interpreting this effect in light of the negative interaction with word count. The positive effect would indicate that words with a higher neighbourhood density are recognised better, however the interaction indicates this effect decreases with higher word count and might be negative for the most frequent words.

% at end of doc for review

\begin{table}
    \caption{Estimated model effects for the gating GLMM and the results of Type III Wald $\chi^2$ tests.}
    \centering
    \resizebox{\linewidth}{!}{
        \begin{tabular}{l r r r r}
        \hline
        Effect & Estimate & Std. error & $\chi^2$ & p \\ \hline
        Intercept & -0.71 & 0.24 & 9.10 & \bf{0.003} \\
        Lemma count & 0.87 & 0.20 & 18.1 & $\mathbf{<0.001}$\\
        Word count & 0.06 & 0.14 & 0.17 & 0.68\\
        \#Vowels & -0.08 & 0.29 & 0.07 & 0.79\\
        \#Consonants & 0.57 & 0.21 & 7.42 & \bf{0.006}\\
        Density & 0.51 & 0.20 & 6.60 & \bf{0.01}\\
        Gate & 0.25 & 0.08 & 11.13 & $\mathbf{<0.001}$\\
        Initial cohort & -0.98 & 0.20 & 23.0 & $\mathbf{<0.001}$\\
        Morphology & -0.02 & 0.23 & 0.01 & 0.92\\
        VQ & -0.09 & 0.05 & 3.18 & 0.07\\
        Speaker id & 0.21 & 0.14 & 2.36 & 0.12\\
        Lemma count:density & 0.19 & 0.13 & 2.09 & 0.15\\
        Word count:density & -0.20 & 0.10 & 4.09 & \bf{0.04}\\
        VQ:gate & 0.03 & 0.01 & 11.61 & $\mathbf{<0.001}$\\\hline
        \end{tabular}
    \label{gating_glmm}
    }   
\end{table}

\subsection{Plurality}\label{plur}

Using the plurality annotations of the visual referents for the noun target words, we test whether the VGS models actually learn to differentiate between singular and plural nouns. That is, if we present it with a plural noun, does it return pictures with multiple visual referents? For this we first select only those target words which have both a plural and singular form. Then, we only keep those words which have at least ten images depicting a single visual referent and ten images with multiple visual referents. So, in theory the VGS models can achieve a perfect P@10 score on these words while also perfectly distinguishing between singular and plural nouns. This results in a final target word set of 28 nouns.  

Table \ref{plurality} shows the confusion matrices for the LSTM and LSTM-VQ models. This shows the number of images containing a single or multiple visuals referents returned when the model is presented with a singular or a plural target word. We see that both VGS models return predominantly images with a single visual referent when presented with singular nouns. When presented with plural nouns, both models return a larger proportion of images with multiple visual referents then when presented with a singular noun (LSTM: $\chi^2(1)=49.8, p<0.0001, N=11150$, LSTM-VQ: $\chi^2(1)=48.1, \\p<0.0001, N=10520$). 

% at end of doc for review

\begin{table}
    \caption{Confusion matrices for singular and plural nouns indicating how many of the correctly retrieved images contained only one or multiple visual referents to the target word.}
    \centering
        \begin{tabular}{l| l l l }

            &\multicolumn{3}{c}{Noun morphology} \\ \hline
            LSTM &\\
            \multirowcell{3}{\\No. referents\\ in image} & & singular & plural\\
            & one & 3048 (57\%) & 2940 (51\%)\\
            & multiple & 2281 (43\%) & 2881 (49\%) \\ \hline
             LSTM-VQ &\\
            \multirowcell{3}{\\No. referents\\ in image} & & singular & plural  \\
            & one & 2857 (56\%) & 2631 (49\%)\\
            & multiple & 2278 (44\%) & 2754 (51\%) \\ \hline
            
        \end{tabular}
    \label{plurality}
\end{table}

\begin{figure*}
    \centering
    \includegraphics[width =.90\linewidth]{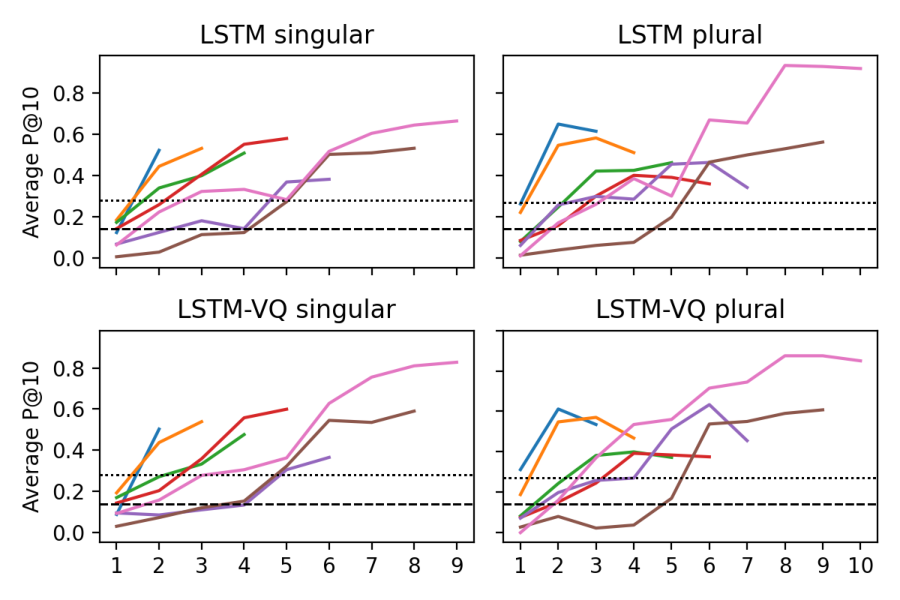}
    \caption{Plots of the recognition scores as a function of the gate number (the number of phonemes processed so far). The solid lines represent averaged P@10 scores over words with an equal number of phonemes (the length of each line indicates the number of phonemes). The dotted and dashed lines represent the naive and random baseline scores respectively.}
    \label{gating_lineplot}
\end{figure*}

Crucially, recognition of plural nouns should depend on the plural suffix, as this is mainly what allows the model to discern whether a target word is plural or singular (although more subtle prosodic cues might also be at play \cite{kemps}). Figure \ref{gating_lineplot} shows the P@10 scores from Experiment 2 as a function of the gate number (phonemes processed so far). We averaged the P@10 scores over words of the same phoneme length. Unsurprisingly, recognition scores tend to increase as more phonemes are processed. Interestingly, for the plural nouns, recognition scores tend to drop at the last phoneme which, except for `men' and `women', is the plural suffix /z/ or /s/. The average P@10 value for plural target words drops from .517 to .479 between the penultimate and final gate for the LSTM model and from .513 to .449 for the LSTM-VQ model. It seems both VGS models have difficulty processing this suffix, the LSTM-VQ model even more so than the LSTM model. 

Nevertheless, as shown in Table \ref{plurality}, the model is able to correctly differentiate between singular and plural nouns, which would indicate that the model has learned to correctly process plural suffixes. A possible explanation for the P@10 drop is that the plural suffix causes the model to retrieve fewer images with single visual referents and more images with multiple referents (as expected) with the drop in single referent images greater than the increase in multiple referent images.

Table \ref{plurality_gating} shows the same confusion matrices as in Table \ref{plurality} but for the phoneme sequence up to the penultimate gate instead of the full word. The numbers between brackets indicate the difference with having seen the full target word. As expected, for singular nouns retrieval of both single and multiple referent images is lower, as the target word is incomplete. For the plural nouns, the model now retrieves more images with a single visual than after having seen the full word while the number of images with multiple visual referents is lower, as the target word is at this point missing its plural suffix. This means that, as hypothesised above, seeing the plural suffix causes a drop in retrieval of single referent images that is greater than the increase in multiple referent images, explaining the drop in P@10 in Figure \ref{gating_lineplot}.

% at end of doc for review

\begin{table}
    \caption{Confusion matrices for singular and plural nouns indicating how many of the correctly retrieved images contained only one or multiple visual referents to the target word. Here we show the scores at the penultimate phoneme and indicate the difference with the scores for the full word between brackets.}
    \centering
        \begin{tabular}{l| l l l }

            &\multicolumn{3}{c}{Noun morphology} \\ \hline
            LSTM &\\
            \multirowcell{3}{\\No. referents\\ in image} & & singular & plural\\
            & one & 2470 ($-$578) & 3339 (399)\\
            & multiple & 1851 ($-$430) & 2694 ($-$187) \\ \hline
             LSTM-VQ &\\
            \multirowcell{3}{\\No. referents\\ in image} & & singular & plural  \\
            & one & 2374 ($-$483) & 3171 (540)\\
            & multiple & 1704 ($-$574) & 2565 ($-$189) \\  \hline
            
        \end{tabular}
    \label{plurality_gating}
\end{table}

\section{Discussion}

In this study we investigated the recognition of isolated nouns and verbs in a Visually Grounded Speech model. Importantly, we are interested in whether the visual grounding allows the model to learn to recognise words as coherent linguistic units, even though our model is trained on full sentences and at no point receives explicit information on word boundaries or that words even exist at all. \cite{havard2019word} used synthetic speech to test word recognition in their VGS model, we used newly recorded real speech. We could have opted to extract the words from spoken captions in the test set but this has a few disadvantages. Firstly, words in a sentence context are often significantly reduced. It has been shown that human listeners have difficulty recognising reduced word forms in isolation even though they are perfectly recognisable in their original sentence context \cite{Ernestus2002}. Secondly, due to co-articulation, we would need to either further reduce the words by removing affected phones, or leave in such information, so that we are not really testing for single-word recognition.

\subsection{Word recognition}

Our first goal was to investigate whether the VGS model learns to recognise words in isolation after only having seen them in a sentence context. Our word recognition results show that our VGS model is able to recognise isolated target nouns. We have even shown that the LSTM model recognises both plural and singular nouns equally well even though plural word forms occur less often in the training data than singular forms. While our scores are lower than those reported in \cite{havard2019word}, some difference was to be expected when working on real as opposed to synthetic speech. The average P@10 scores indicate that more than half of the top 10 retrieved images contain the correct visual referent and the model scores well above the baselines. In fact, only four words (two in the LSTM model and two in the LSTM-VQ model) were not recognised at all, namely `river' (in both models), `ball' (LSTM) and `waves' (LSTM-VQ). We saw that `river' does however return pictures with other bodies of water in it (e.g., lakes or the ocean), and indeed it can be hard to discern the difference between a lake and a river from a picture. The fact that `ball' is not recognised is a little baffling considering that `basketball' has a P@10 score of .8 and `football' a score of .4 (and pictures of either are also annotated as ordinary balls).

We also tested whether the model is able to recognise verbs. While we tested verbs in root, third person and participle form, the participle form is most common in the image descriptions. But even when we look only at the scores on the participle form, recognition scores for verbs are much lower than for nouns. In fact, most words are not recognised at all, and only 11 (LSTM) or 12 (LSTM-VQ) words had P@10 scores over .5. Looking at these words we see that many of them would consistently occur together with an object (e.g. `surfing', `playing', `skiing', `holding' and `racing'), so the model might simply recognise the objects they co-occur with. This could be explained by our use of image features from a network trained to recognise objects, not actions or body postures. However, the model also recognises `running', `walking', `jumping' and `smiling', so the image features do seem to contain more information than simply the presence of a human in the image. Verb recognition in our model was far from good and this presents an interesting avenue for further research. We think it is possible for the VGS model to also learn to recognise actions, perhaps by fine-tuning parts of ResNet with the VGS model or training the visual side of the model from scratch like in \cite{harwath2018jointly}. 

\subsection{Word competition}

In our gating experiment, we investigated whether the model’s word recognition is affected by word competition, as is the case in humans. The results of this experiment show clear evidence of word competition effects in our model. There is a strong negative effect of word-initial cohort where recognition scores are lower as more words are possible given the current input sequence. We also find a positive effect of neighbourhood density which is modulated by a negative interaction with word count. This means that the effect of neighbourhood density is higher for low-frequency words and vice versa. This is in line with the findings of \cite{metsala,goh} that for humans recognition of low-frequency words is facilitated by dense neighbourhoods and recognition of high-frequency words is facilitated by sparse neighbourhoods. 

However, we find a positive main effect of neighbourhood density, contrary to what we may expect given that denser neighbourhoods lead to more word competition. Furthermore, given the strength of the interaction with word count, the neighbourhood density effect would only be negative at high levels of (log) word count. \cite{metsala} gives a possible explanation for the interaction between word count and neighbourhood density. During word learning, dense neighbourhoods have a positive effect on word recognition; hearing similarly sounding words facilitates learning. During word recognition, dense neighbourhoods have a negative effect; similarly sounding words compete for recognition. For infrequent words, the learning effect outweighs the competition effect, and vice versa. Our model may simply have seen too few of the most frequent words for the competition effect to outweigh the learning effect, explaining the positive effect of neighbourhood density. Together with the strong effect of initial cohort, we argue that we do indeed see word competition effects in our VGS model.

\subsection{Plurality}

We also investigated whether our VGS model learns the difference between singular and plural nouns. Our results show that not only is the model able to recognise target nouns in both forms but, to a limited extent, it also learns to differentiate between the two forms. When prompted with plural target nouns, the model retrieves more images with multiple referents and fewer with single referents than when prompted with single nouns, as is to be expected if a model differentiates between singular and plural nouns (see Table \ref{plurality}). Thus, the model learns a meaningful difference between singular and plural nouns in terms of their visual representations. 

P@10 scores from our gating experiment showed that words are recognised better as more of the word is processed, as was to be expected. Yet, we also see that recognition scores are well above zero and even above the baselines before word offset, which means that the model is able to recognise words from partial input. We take this to mean that the model not only learns to recognise words, but is also able to encode useful sub-lexical information. However, both models seemed at first glance to have trouble with the plural suffix. 

As shown by the results of the gating experiment, recognition of plural target words is often higher than recognition of singular target words before the plural suffix, but drops at the final phoneme, at which point recognition scores of plural nouns is equal to singular nouns for the LSTM model and lower for the LSTM-VQ model. While this seems to be evidence against the encoding of useful sub-lexical information, our results also show that presenting the model with plural nouns causes both models to retrieve \emph{more} images with multiple visual referents and \emph{fewer} images with a single referent. This indicates that the model encodes the plural suffix in a way that correctly affects recognition. 

Using the recognition results from the gating experiment, we tested whether it was indeed only after the plural suffix was processed that the distribution of single or multiple referents in the retrieved images shifts, and we found this to be the case. At the gate just before the plural suffix (where the word is technically still singular), the model retrieves more single referent images and less multiple referent images than after having seen the plural suffix. As previously said this is in contrast to human listeners, who are able to use more subtle prosodic cues to recognise plural nouns \cite{kemps}. It is not surprising that our current model, which is far from human performance in terms of word learning and recognition, is not able to exploit such cues, but this is an interesting avenue for further research. 

Further analysis showed that after processing the plural suffix, the drop in single referent images is larger than the increase in multiple referent images. This may simply be caused by an imbalance in the test data; there are more annotations of single visual referents (3,864) than multiple visual referents (2,203). Further testing with a more balanced set of test images could show whether the performance drop seen in our gating experiment is indeed due to \emph{correct} recognition of the plural suffix, as we would then expect a balanced increase and decrease in the number of single and multiple referent images that are retrieved. 

\subsection{Vector Quantisation}

Our final research goal was to establish whether the addition of VQ layers to the VGS model aids in the discovery and recognition of words. Previous research had shown that VQ layers inserted into a VGS model learned a hierarchy of linguistic units; a phoneme-like inventory in the first layer, and a word-like inventory in the second layer \cite{HarwathVQ}. VQ layers discretise otherwise continuous hidden representations and furthermore learns to map neighbouring speech frames to the same embedding in the codebook. We expected that this aids in the discovery of words and perhaps even allows the LSTM-VQ model to recognise words earlier in the gating experiment, as the model is forced to output discrete units from its word-like VQ layer at every time step. Moreover, the codebook size (2048) is smaller than the total number of unique words in Flickr8k so, if anything, one would expect the model to prioritise highly frequent words, of which we took the top 50 as our targets. 

In all of the experiments, however, we found no evidence of the VQ layers aiding in the recognition of words: we showed that the LSTM-VQ model performance on the training task (Image-Caption retrieval) is similar to the LSTM model and slightly outperforms it so it cannot be the case that the LSTM-VQ model is simply not a good VGS model. With regard to word recognition performance, the LSTM-VQ model recognises singular nouns better than the LSTM model, but it performs much worse at recognising plural nouns. Also noticeable is a gap between recognition on singular nouns versus plural nouns that is not present in the LSTM model (when looking at the same subset of words that have both a plural and singular form). 

Furthermore, both GLMMs showed no main effect of the presence of VQ layers on recognition scores. 
We did find a positive interaction between VQ and gate number. The positive interaction between VQ and gate indicates that the effect of gate is larger for the LSTM-VQ model than for the LSTM model. This implies that, for early gates, the LSTM-VQ model performs worse than the LSTM model. That is, the LSTM-VQ model recognises words \emph{later} rather than earlier as we expected. Together these results indicate that the addition of VQ layers is neither beneficial nor detrimental to word recognition performance, however the LSTM-VQ model requires more of the input sequence for correct recognition. An interesting question for future research is which model performs more 'human-like', that is, which model recognises words closest to the point where humans are able to recognise them. 

Finally, we did a post-hoc test for the interaction between VQ and word form  that shows the LSTM-VQ model has an advantage on the most frequent noun and verb forms, but performs worse on the other, less frequent, forms. Perhaps this is due to the limited codebook size forcing the model to dedicate codes to the most frequent words in the training data. A possible explanation might be that the codebook only dedicates codes in its limited codebook to the most frequent words. 

\section{Conclusion}

We investigated whether VGS models learn to discover and recognise words from natural speech. Our results show that our model learns to recognise nouns. To a lesser extent, the model is capable of recognising verbs but future research should look to the image recognition side of the model to further improve this. Our model even learned to encode meaningful sub-lexical information, enabling it to interpret the visual difference signalled by the plural morphology. Contrary to what we expected based on previous research, our results show no evidence that vector quantisation aids in the discovery and recognition of words in speech. Importantly, we investigated the cognitive plausibility of the model by testing whether word competition influences our model's word recognition performance, as we know happens in humans. We have shown that two well known measures of word competition predict word recognition in our model and found evidence in favour of a disputed interaction between word count and neighbourhood density found in human word recognition. 

Taking inspiration from human learning processes, our research has shown that using multiple streams of sensory information allows our model to discover and recognise words without any prior linguistic information from a relatively small dataset of scenes and spoken descriptions. We think that using realistic and naturally occurring input is important in order to create speech recognition models that are more cognitively plausible and visual grounding is an important step in that direction.

\section*{Acknowledgements}

Funding: The research presented here was funded by the Netherlands Organisation for Scientific Research (NWO) Gravitation Grant 024.001.006 to the Language in Interaction Consortium. 

\bibliographystyle{vancouver}

\bibliography{mybib}

\end{document}